\documentclass{article}
\usepackage{iclr2027_conference,times}
\iclrfinalcopy

\usepackage{amsmath,amsfonts,bm}

\def\eqref#1{equation~\ref{#1}}

\def\1{\bm{1}}

\DeclareMathAlphabet{\mathsfit}{\encodingdefault}{\sfdefault}{m}{sl}
\SetMathAlphabet{\mathsfit}{bold}{\encodingdefault}{\sfdefault}{bx}{n}

\usepackage{amsmath,amssymb,amsthm}
\usepackage{booktabs}
\usepackage{tabularx}
\usepackage{graphicx}
\usepackage{xcolor}
\usepackage{microtype}
\usepackage{enumitem}
\usepackage{float}
\usepackage{wrapfig}
\usepackage{placeins}
\usepackage{tikz}
\usetikzlibrary{arrows.meta,positioning,fit}
\usepackage{hyperref}
\usepackage{url}

\definecolor{evidenceblue}{HTML}{5F7180}%
\definecolor{warningred}{HTML}{B08272}%
\definecolor{MutedGray}{HTML}{8F8B83}%
\definecolor{morandisage}{HTML}{7E9070}%
\hypersetup{
  colorlinks=true,
  linkcolor=evidenceblue,
  citecolor=evidenceblue,
  urlcolor=evidenceblue
}

\newcommand{\ourtrace}{\textsc{Our Trace}}
\newcommand{\ourfix}{\textsc{Our Fix}}

\newcommand{\taskrecovery}{\mbox{\textsc{TaskRecovery}}}
\newcommand{\localresolve}{\mbox{\textsc{LocalResolve}}}
\newcommand{\verifiedroot}{\mbox{\textsc{VerifiedRoot}}}

\title{\centering From Trace to Fix: Residual Task State\\ in Long-Horizon Agents}

\author{
    Zile Huang \quad Ser-nam Lim \\
    University of Central Florida \\
    \texttt{zi057775@ucf.edu} \quad \texttt{sernam@ucf.edu}
}

\date{}

\begin{document}

\maketitle

\begin{abstract}
Recovering from failures in long-horizon, stateful language agents is challenging because the action introducing a defect, the symptom exposing it, and the persistent state requiring repair frequently diverge. 
Existing methods largely focus on retrospective failure attribution, leaving sequential diagnosis and state repair poorly integrated under constrained budgets. 
We introduce \textbf{TRACE--FIX}, a unified recovery framework centered on versioned residual task state. 
\textsc{Trace} transforms execution logs into a structured dependency-linked ledger, prioritizing unresolved task requirements rather than salient surface errors. 
\textsc{Fix} formulates recovery as budgeted planning over admissible actions, incorporating outcome forecasting and explicit repair-reserve constraints to prevent premature budget exhaustion. 
To rigorously decouple localization accuracy from downstream repair capability, we present \textbf{LongH-Debug}, evaluating root-cause attribution, verified-fault recovery, and end-to-end closed-loop execution independently. 
Our empirical analysis demonstrates that structured residual tracking substantially improves causal localization over raw-trace baselines, while budget-aware sequential planning converts verified fault signals into more reliable task recovery without destructive trial-and-error.
Code is available at \url{https://github.com/toffeecat/tracedp}.
\end{abstract}

\section{Introduction}
\label{sec:introduction}

Language-model agents increasingly complete tasks by changing persistent state:
they write files, update databases, invoke tools, create intermediate artifacts,
and make decisions that later steps depend on
\citep{dong2026longhorizon,zhou2023webarena,xie2024osworld,jimenez2024swebench}.
In such tasks, the action that introduces a problem need not coincide with the
step where the problem becomes visible. The environment can also change again
before debugging begins. As a result, the historical cause, the observed
symptom, and the state that currently needs intervention can be different.

A retained database-recovery run illustrates the distinction. A successful
SQLite operation silently removes the write-ahead log that contains records
needed later in the task. Subsequent commands continue normally, the actor
eventually recognizes the mechanism, and the final failure occurs long after
the destructive state transition. At recovery time, replaying the final error
or blaming the latest suspicious command does not identify what information
was lost; restarting from scratch would also discard work that remains valid.
The useful recovery question is instead: \emph{which task requirement is still
unsupported, which historical state change is relevant to it, and what
affordable action can improve the task from here?}

\newcommand{\stepok}{{\color{morandisage}$\checkmark$}}
\newcommand{\stepbad}{{\color{warningred}$\times$}}
\newcommand{\steprun}{{\color{MutedGray}$\cdot$}}
\newcommand{\modehead}[3]{%
  {\small\bfseries #1}\hfill{\scriptsize\itshape\textcolor{MutedGray}{#2}}\par
  \vspace{1pt}{\scriptsize\sffamily\textcolor{evidenceblue}{#3}}\par
  \vspace{1.5pt}\textcolor{evidenceblue}{\rule{\linewidth}{0.6pt}}\par\vspace{2pt}}
\newcommand{\stepline}[2]{{\scriptsize #1}\hspace{0.3em}{\ttfamily\scriptsize #2}\par\vspace{0.8pt}}
\newcommand{\modenote}[1]{\vspace{2pt}{\scriptsize #1}\par}

\begin{figure}[t]
\centering
\setlength{\tabcolsep}{0pt}
\begin{tabular}{@{}p{0.48\linewidth}@{\hspace{0.04\linewidth}}p{0.48\linewidth}@{}}

\begin{minipage}[t]{\linewidth}
\modehead{A\quad Delayed state mutation}{db-wal-recovery}{live run, 97 tool calls}
\stepline{0}{\steprun\ ls -la /app: base 8192\,B $+$ log 16512\,B}
\stepline{2}{\stepbad\ sqlite3 main.db\ \dots\ PRAGMA wal\_checkpoint: log deleted}
\stepline{9}{\steprun\ cp main.db-wal /tmp/\dots bak\ \ (already too late)}
\stepline{38}{\stepbad\ the same read deletes the rebuilt log again}
\stepline{95}{\steprun\ the actor states the mechanism itself}
\stepline{end}{\stepbad\ timeout, reward 0.0}
\modenote{A successful read removes the only carrier of the six missing records. Nothing errors, so the run repeatedly rebuilds and destroys the log.}
\end{minipage} &

\begin{minipage}[t]{\linewidth}
\modehead{B\quad Residual obligation}{openssl-selfsigned-cert}{live run, 7 checks}
\stepline{1}{\stepbad\ script-runs: /app/check\_cert.py exits 1}
\stepline{2}{\stepok\ verification-file: subject and fingerprint}
\stepline{3}{\stepok\ key-attributes: 2048-bit RSA, mode 600}
\stepline{4}{\stepok\ script-runs re-run: the same predicate passes}
\stepline{5}{\stepok\ combined-pem parses as a certificate}
\stepline{6}{\stepok\ ssl-directory-layout holds the deliverables}
\stepline{7}{\stepok\ certificate-attributes: subject, 365-day validity}
\stepline{end}{\stepok\ reward 1.0}
\modenote{The actor rewrites the script after the first failure; the same predicate then passes and the task reaches reward $1.0$.}
\end{minipage} \\

\addlinespace[8pt]

\begin{minipage}[t]{\linewidth}
\modehead{C\quad Blame is not a current fault}{bn-fit-modify}{live run, 8 checks}
\stepline{0}{\stepok\ learned\_dag.csv exists in to,from format}
\stepline{1}{\stepok\ intervened DAG exists in to,from format}
\stepline{2}{\stepok\ intervened sample keeps the column names}
\stepline{3}{\stepok\ the DAG has six total edges}
\stepline{4}{\stepok\ intervened edges are a subset of the learned DAG}
\stepline{5--7}{\stepok\ repeated checks of the same conditions}
\stepline{end}{\stepok\ 8/8 supported, reward 1.0, 0 repairs}
\modenote{The history contains a failure, but the live state already satisfies every check. Historical blame is therefore not a current repair target.}
\end{minipage} &

\begin{minipage}[t]{\linewidth}
\modehead{D\quad Budgeted recovery}{cancel-async-tasks}{live run, 8 checks, 2 repairs}
\stepline{1--2}{\stepok\ import-form and signature supported}
\stepline{3}{\stepbad\ concurrent-execution refuted}
\stepline{4--6}{\stepok\ max-concurrency, cancel-below, cancel-at supported}
\stepline{7}{\stepbad\ cancel-above-limit refuted}
\stepline{8}{\stepok\ import-form re-run: supported}
\stepline{end}{\stepbad\ 2 repairs tried, reward 0.0}
\modenote{Eight checks and two repairs exhaust the twelve-unit budget before either repaired predicate is re-verified.}
\end{minipage} \\

\end{tabular}
\caption{\textbf{Four long-horizon failure modes from retained Terminal-Bench trajectories.}
(A) A successful operation silently destroys state whose loss appears only later.
(B) A failed predicate is subsequently repaired, leaving only its residual consequences relevant.
(C) A historical failure remains in the trajectory although the current state is already healthy.
(D) Checks and repairs share a finite budget, which can be exhausted before repair verification.
Rows quote retained commands or registered checks; \stepok\ and \stepbad\ denote supported and refuted outcomes.}
\label{fig:failure-modes}
\end{figure}

Prior work has made substantial progress on failure attribution, critical-step
localization, counterfactual replay, and intervention-based verification
\citep{zhang2025whowhen,chen2026traceelephant,ma2025dover,
liu2026whowhenpro,zhang2025agentracer,qiao2026verifymas}.
These methods primarily organize evidence about \emph{what went wrong}.
Recovery adds a sequential decision problem: additional diagnosis consumes
time and intervention budget, repair can invalidate earlier evidence, and the
best next action depends on what will remain possible after its outcome.
A more complete diagnosis is therefore not automatically a better recovery
decision.

We address this problem with two components over a shared runtime state.
\textbf{TRACE} represents current task requirements together with versioned
execution evidence and ranks historical events that can explain requirements
that remain unresolved. \textbf{FIX} plans over the legal checks, recovery
actions, actor continuation, and submission choices available at the current
boundary. Its short-horizon outcome estimates are supplied from public runtime
state, but action identity, costs, state transitions, and the dynamic-programming
backup are enforced by runtime code. Every executed action produces a receipt;
resource changes invalidate version-bound evidence, the residual state is
recompiled, and the next decision is made from the new state.

We also introduce \textbf{LongH-Debug}, which separates retrospective
localization from executable recovery. Localization is evaluated under matched
public inputs; recovery methods are compared under the same independently
verified fault packet; and a final factorial crossing tests whether the
localization and recovery components compose end to end. This separation avoids
crediting a fixer for hidden re-diagnosis or penalizing it for an upstream
localization error.

\paragraph{Contributions.}
First, we define a versioned residual task representation that distinguishes
current unsupported requirements from historical events that no longer require
intervention. Second, we connect this representation to a practical recovery
controller that combines public-state outcome forecasting with runtime-checked
bounded planning and replanning after real execution. Third, LongH-Debug
provides matched localization, verified-fault recovery, and closed-loop
evaluation protocols so that attribution quality and recovery quality can be
measured separately before they are composed.

\section{Method}
\label{sec:method}

\begin{figure}[!t]
    \centering
    \includegraphics[width=\linewidth]{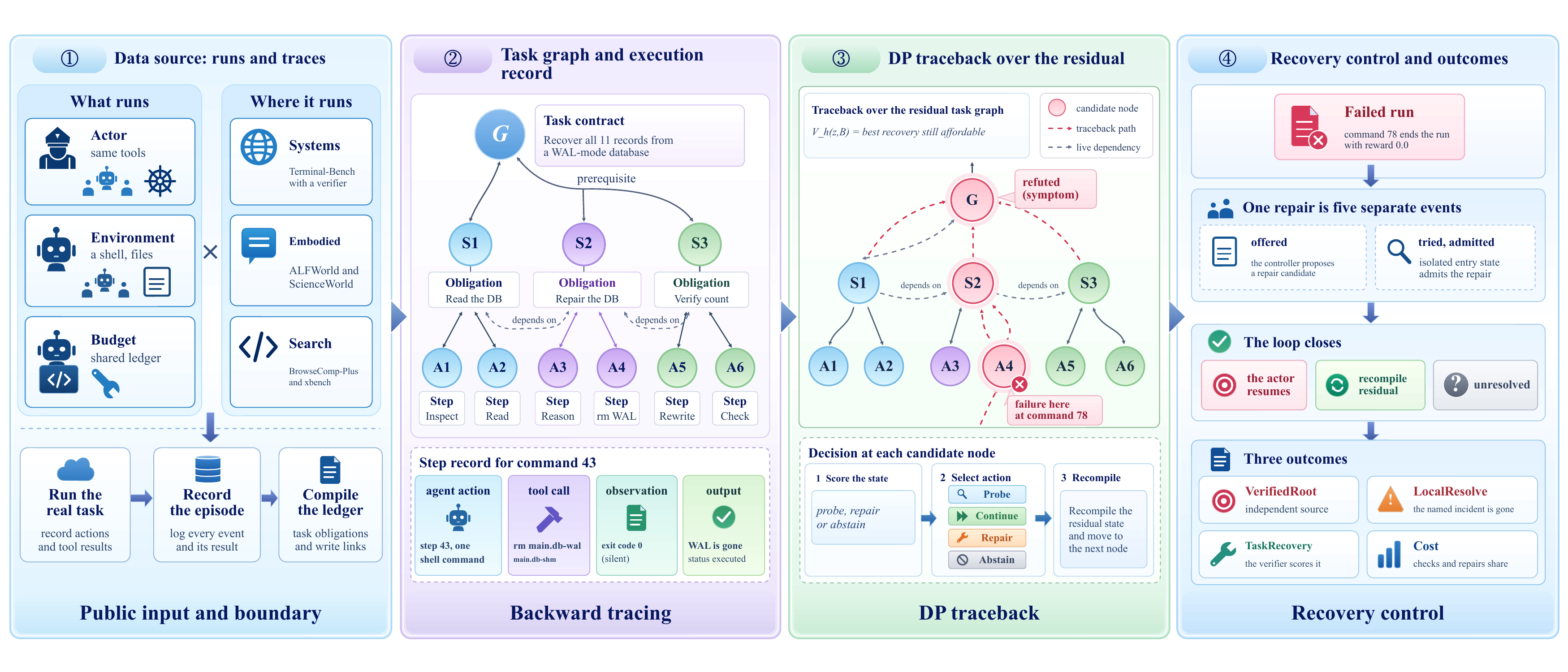}
    \caption{\textbf{Runtime path from public history to recovery.}
    The task contract, public execution record, resource versions, and registered
    checks are compiled into a residual state. TRACE uses the resulting
    dependency-linked ledger to rank historical events relevant to unresolved
    requirements. FIX receives the live action slate and a bounded public-state
    forecast, while runtime code checks action identity, cost, reachability, and
    time before computing the depth-two decision. The selected action is then
    executed; its receipt updates resource versions and evidence, the residual
    state is recompiled, and planning repeats before the same actor continues.
    Hypothetical planning outcomes are kept separate from observed execution
    receipts throughout.}
    \label{fig:overview}
\end{figure}

Figure~\ref{fig:overview} summarizes the runtime path. The key design choice is
to treat the execution history as evidence used to construct the current
recovery state, rather than as the state itself.

\subsection{Versioned residual task state}
\label{sec:residual-state}

The runtime maintains three linked views with different semantics.
The \emph{task view} contains public requirements, their prerequisite
relationships, and required outputs. The \emph{execution ledger} contains
ordered actions, observations, resource interactions, and resource versions.
The \emph{acceptance view} contains task predicates together with the evidence
and incidents that currently support or refute them. These references make a
historical event relevant to a task requirement without treating a dependency
or last-writer edge as causal proof
\citep{weiser1981slicing,ferrante1987pdg,agrawal1990dynamic}.

Evidence is version-bound. When evidence is recorded for a proposition, the
runtime stores the versions of the resources on which that evidence depends.
If a later observation changes any of those resources, the evidence is marked
\emph{stale}; task revisions supported only by stale evidence are excluded when
the live task view is rebuilt. We call a requirement a \emph{residual
obligation} when it is not currently supported, is blocked by an unresolved
prerequisite, or has lost its support through such a version change.

The recovery state consists of these residual obligations, their current
evidence, the historical events linked to them, and the legal action slate.
We keep the remaining intervention budget separate from this state
representation; a planner decision is made from the compiled state together
with the remaining budget. This avoids counting budget both inside a state
transition and again in the planning recursion.

\subsection{TRACE: structured history, explicit ranking}
\label{sec:trace}

TRACE first compiles the public execution into a residual ledger. Each event
retains its identity and order and is annotated mechanically with its actor or
tool summary, observed result, referenced resources, explicit write targets,
and upstream writer links. The task contract and current residual obligations
provide the context in which these event-level effects are interpreted.

The current ranking stage is explicit rather than implicit in the graph
construction. An audited LLM call receives the public task specification,
the observed failure, and the compiled residual ledger and returns the top
historical event IDs, ranked from most to least likely to have created an
unresolved task condition. The prompt asks the ranker to distinguish
state-changing writers from later observations, tests, or reports of the
failure, and to consider omitted or misordered actions when no explicit harmful
write explains the outcome. The ledger therefore structures the evidence
presented to the ranker; it does not deterministically declare the root cause.

TRACE writes its result to an immutable \emph{fault packet} containing the
ranked event hypotheses, their public evidence references, the residual
obligations they may affect, and qualified checks that can test them.
Verification remains separate from ranking: a high-ranked event is a
localization hypothesis until an admissible check, replay, or intervention
supports it.

\subsection{FIX: bounded planning over the live action slate}
\label{sec:fix}

FIX does not operate over a fixed four-action alphabet. At each recovery
boundary the runtime emits the actions that are actually legal for that state,
with stable action identities, target bindings, and declared costs. Depending
on the environment, the slate can include actor continuation, registered
checks, task- or resource-targeted recovery requests, additional diagnosis or
completion requests, and submission of the current state.

The current planner uses a separate no-tool LLM as a \emph{forecast model}.
It receives only public runtime state: the task graph, residual predicate
states, registered checks, a bounded window of recent actor execution
summaries, the live candidate catalogue, remaining time, and remaining
intervention budget. For each root action it predicts bounded numeric outcomes;
for checks these include masses for \emph{supported}, \emph{refuted}, and
\emph{unknown} together with estimates of what the same actor can accomplish
after each verdict. It may additionally expand at most two reachable
second-step actions. Terminal leaves carry a compact estimate of task success
and the fraction of the residual problem expected to remain unblocked.

These forecasts are planning estimates, not calibrated probabilities.
Runtime code validates their schema and numeric ranges, derives outcome
semantics from the action type, rejects transitions that create time or budget,
and compiles the accepted forecast into a depth-two decision tree. Standard
dynamic programming then compares the expected downstream recovery of the
affordable candidates. The model never outputs the selected action or a
Bellman value directly; both are recomputed by the runtime from the admitted
forecast. If the forecast is missing or fails an admission gate, the runtime
falls back to actor continuation rather than inventing a transition.

The intervention budget is an abstract, frozen cost ledger rather than a token
or dollar measure. In the evaluation contract, each declared TRACE,
FIX-controller, or actor-continuation model call costs one unit; executable
checks and repairs use the costs frozen in the environment's qualification
catalogue. Tokens, wall-clock time, and monetary cost are reported separately.
This shared ledger makes ``inspect more'' and ``preserve budget for repair''
directly comparable.

\subsection{Execution, recompilation, and handoff}
\label{sec:closed-loop}

Planning ends when an action is selected; recovery evidence begins only after
that action is executed. The execution receipt records the observed result and
the resource versions at the resulting boundary. Any changed resource
invalidates evidence bound to its previous version, incidents and predicate
states are updated, and the residual task is recompiled before another
decision is allowed.

A local intervention is not treated as task success. After its postcondition is
checked, control returns to the same actor when continuation is legal; the
benchmark's final verifier scores the state produced by that continuation.

\section{LongH-Debug Benchmark}
\label{sec:benchmark}

\begin{figure}[t]
    \centering
    \includegraphics[width=\linewidth]{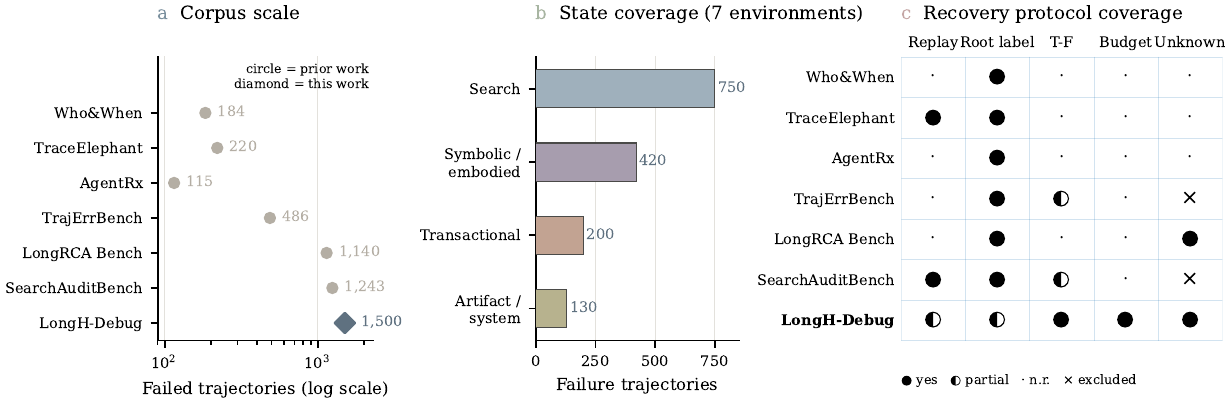}
    \caption{\textbf{LongH-Debug benchmark: scale, state diversity, and
    executable recovery coverage.}
    (a) Published failure-trajectory counts for prior agent-debugging benchmarks,
    with the LongH-Debug row shown separately.
    (b) The planned 1,500-trajectory corpus allocates quotas across seven environments grouped into
    search/retrieval, symbolic/embodied, transactional/social, and persistent
    artifact/system state.
    (c) Protocol coverage distinguishes retrospective attribution from executable
    recovery. Filled circles denote explicitly supported capabilities, half-filled
    circles partial or alternative protocols, dots unreported capabilities, and
    crosses explicit exclusions. LongH-Debug's replay and labels apply to its
    qualified subsets rather than to every trajectory in the planned corpus.}
    \label{fig:corpus-scale}
\end{figure}

LongH-Debug is organized in two tiers so that corpus scale is not confused with
recovery capability. A \emph{failure record} contains a public task
specification, the actor-visible trajectory, and the public failure signal
needed for localization. A \emph{recovery-qualified case} additionally requires
a reproducible failed entry state, the task/resource/version records needed to
reconstruct residual obligations, a frozen catalogue of executable checks and
repairs, a common cost ledger, and an independent final verifier. Only this
qualified subset is used for experiments that execute recovery actions.

\paragraph{Source environments.}
The corpus draws natural failures from seven environments with different state
mechanisms. Terminal-Bench~\citep{merrill2026terminalbench} supplies persistent
filesystem and artifact state; $\tau^2$-bench~\citep{yao2024taubench} supplies
transactional state across tool-mediated conversations; ALFWorld
\citep{shridhar2021alfworld}, ScienceWorld~\citep{wang2022scienceworld}, and
TextWorld~\citep{cote2019textworld} supply symbolic embodied state; and
BrowseComp-Plus~\citep{chen2025browsecompplus} and xbench~\citep{chen2025xbench}
supply long search trajectories. We preserve each environment's native task
semantics rather than converting them into a shared synthetic state space.

We plan to collect $1{,}500$ natural failure trajectories, with each
environment quota bounded by the executions that the source environment can
actually provide. The quotas are planned allocations, not counts of collected
trajectories.

\paragraph{Long-horizon annotations.}
We do not use transcript length as the definition of a long-horizon failure.
For localization cases with supported dependency annotations, we record two
separate axes. \emph{Dependency depth} $H_c$ counts the supported dependency
chain connecting a decisive historical event to the residual obligation it
blocks. \emph{Feedback delay} $H_f$ counts the decisions between that event and
the first public evidence that reveals its consequence. The localization
population is stratified by these axes, allowing a long transcript with a
local fault to be distinguished from a failure with genuinely distant
dependencies or delayed evidence.

\paragraph{Evaluation populations and outcomes.}
LongH-Debug defines three matched populations.
TRACE is evaluated on $N=200$ task-disjoint natural failure records under
identical public inputs. FIX is evaluated on $N=120$ recovery-qualified cases
where every fixer receives the same independently verified fault packet. The
full TRACE$\times$FIX crossing uses $N=60$ qualified tasks restored from the
same failed entry state and scored by the same official verifier.

The protocol records three different outcomes. \verifiedroot{} asks whether an
emitted historical hypothesis can be independently supported within the allowed
verification budget. \localresolve{} asks whether the targeted current incident
is gone after intervention and continuation. \taskrecovery{} is the final
benchmark-owned task outcome. These quantities are intentionally not collapsed
into one score: locating a historical cause, removing a current blocker, and
finishing the original task are different events.

Historical annotations and root-verification evidence remain evaluator-side.
For paired FIX comparisons, the admitted fault packet is frozen and shared
unchanged across methods. All launched recovery attempts remain in the
denominator, including budget exhaustion and \emph{unknown} check outcomes.

\begin{table}[t]
\caption{\textbf{TRACE localization on $N=200$ paired natural failures.}
Candidate recall asks whether the adjudicated event is present before ranking;
\verifiedroot@$B$ requires an independent qualified check sequence within the
verification budget. Token counts are total model input tokens over the paired
population and relative cost is normalized to Generic one-shot.}
\label{tab:trace-main-expanded}
\centering
\scriptsize
\setlength{\tabcolsep}{3.2pt}
\renewcommand{\arraystretch}{1.08}
\begin{tabular}{lrrrrrrr}
\toprule
Method & Hit@1 & Hit@3 & MRR & Cand. recall & \verifiedroot@$B$ & Tok. (M) & Rel. cost \\
\midrule
Generic one-shot        & 42.0 & 74.5 & .571 & 93.0 & 31.0 & 2.40 & 1.00 \\
Dependency-only         & 24.0 & 46.0 & .351 & 64.0 & 15.5 & 0.52 & 0.22 \\
Who\&When (adapted)     &  8.0 & 12.0 & .097 & 17.0 &  4.5 & 6.00 & 2.50 \\
TraceElephant (adapted) & 44.0 & 61.0 & .508 & 71.0 & 33.0 & 12.20 & 5.08 \\
Residual Context        & 43.0 & 74.5 & .573 & 93.0 & 32.0 & 7.80 & 3.25 \\
\textbf{\ourtrace{}}    & \textbf{68.0} & \textbf{89.0} & \textbf{.782} &
\textbf{97.0} & \textbf{60.5} & 3.10 & 1.29 \\
\bottomrule
\end{tabular}
\end{table}

\begin{table}[t]
\caption{\textbf{Recovery under matched verified-fault input ($N=120$).}
 \taskrecovery{} is the
benchmark-owned outcome after same-actor continuation; \localresolve{} asks
whether the targeted live incident is gone. Harmful counts unnecessary or
state-damaging interventions. Cost/success is normalized to Same-Actor Continue.}
\label{tab:fix-main-expanded}
\centering
\scriptsize
\setlength{\tabcolsep}{4.0pt}
\renewcommand{\arraystretch}{1.08}
\begin{tabular}{lrrrr}
\toprule
Fixer & \taskrecovery{} & \localresolve{} & Harmful & Cost/success \\
\midrule
Same-Actor Continue             & 35.0 & 29.2 & 0.0 & 1.00 \\
Fresh Fixer                     & 42.5 & 40.8 & 5.8 & 0.92 \\
Fixed-order sweep               & 47.5 & 50.0 & 4.2 & 0.86 \\
Information gain                & 50.0 & 53.3 & 4.2 & 0.82 \\
Information gain + repair reserve & 61.7 & 66.7 & 3.3 & 0.69 \\
Residual-Greedy                 & 56.7 & 61.7 & 3.3 & 0.73 \\
\textbf{\ourfix{} depth-2}      & \textbf{67.5} & \textbf{73.3} & \textbf{2.5} & \textbf{0.61} \\
\bottomrule
\end{tabular}
\end{table}

\section{Experiments}
\label{sec:experiments}

We organize the evaluation around four questions.
\textbf{RQ1} asks whether TRACE localizes historical events that remain relevant
to the current task, rather than merely retrieving a salient failure step.
\textbf{RQ2} gives every fixer the same independently verified fault packet and
asks whether FIX converts identical fault information into better recovery.
\textbf{RQ3} crosses tracing and fixing components on the same failed entry
states to test whether the two gains survive end to end.
\textbf{RQ4} isolates the state, planning, and recompilation mechanisms under
matched information and action sets.

\subsection{Setup and comparison contract}
\label{sec:exp-setup}

The three primary populations are those defined in
Section~\ref{sec:benchmark}: $N=200$ paired localization cases, $N=120$
verified-fault recovery cases, and $N=60$ closed-loop tasks.
Within a comparison cell we freeze the public trajectory or failed entry state,
actor identity, model configuration, legal tools, action catalogue, cost ledger,
and final verifier.

For localization, \emph{Generic one-shot} is a direct causal ranker over the
public trajectory. \emph{Dependency-only} uses the structured dependency slice
without the residual-causal ranking prompt. \emph{Residual Context} gives a
ranker the corresponding residual information as additional flat context.
Who\&When is run through its pinned upstream attribution code after projecting
each public LongH event into one history entry with its public actor/component
label and leaving the upstream ground-truth field empty.
TraceElephant retains its upstream prompt, search, fallback, and parsing code,
while canonical public events are reconstructed into the structured
request/response fields expected by the released implementation; public
evaluator failure is mapped into its test-status input and ground truth remains
empty.

For recovery, all FIX methods receive the same independently verified fault
packet. We compare against \emph{Same-Actor Continue}, a \emph{Fresh Fixer},
a fixed-order check sweep, one-step residual-greedy control, and a strong
information-gain policy with an explicit repair-budget reserve.
The last comparator is important: controlled development experiments found
that repair-budget reservation, not Bellman recursion alone, explained the
observed gain over naive information gain. The formal comparison therefore
does not use a deliberately budget-oblivious policy as the strongest baseline.

\subsection{RQ1: Does TRACE localize the historical event that still matters?}
\label{sec:trace-results}

\begin{figure}[t]
    \centering
    \includegraphics[width=\linewidth]{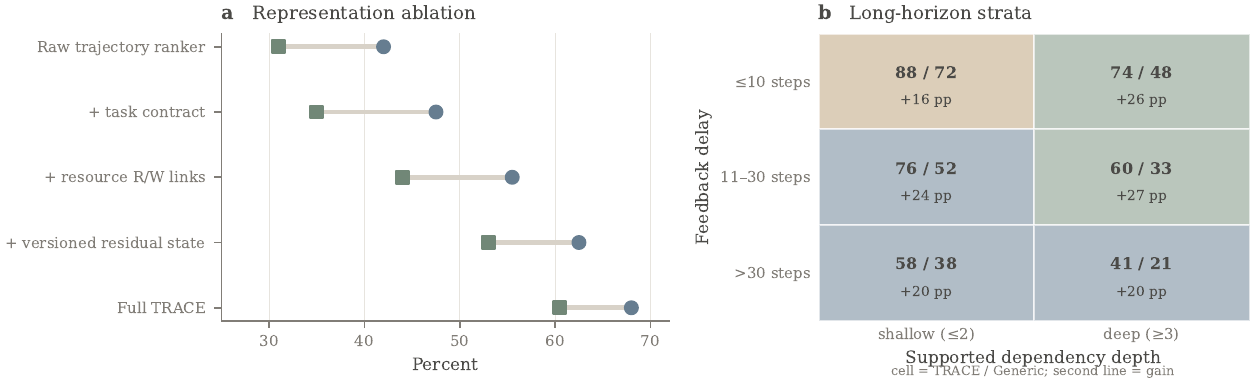}
    \caption{\textbf{Why TRACE should help.}
    (a) Representation ablation. Only the endpoints, Raw trajectory ranker and
    Full TRACE, are measured (Table~\ref{tab:trace-main-expanded}); the three
    intermediate rows are planned targets with no measured values yet.
    (b) TRACE/Generic Hit@1 by supported dependency depth and feedback delay.
    Each cell is a planned target with no measured value yet.
    Circles mark Hit@1 and squares mark VerifiedRoot@B.}
    \label{fig:trace-evidence}
\end{figure}

Table~\ref{tab:trace-main-expanded} evaluates every tracer on identical public
inputs. We report ranking quality, candidate coverage, executable verification,
and inference cost together; a localization method that retrieves the correct
event only through substantially more context should not be read as an
unqualified improvement.
The headline is not just a Hit@1 gain. TRACE is intended to preserve
high candidate recall while converting that coverage into a substantially
higher fraction of hypotheses that can be independently verified inside the
same downstream budget. Figure~\ref{fig:trace-evidence} separates the
representation contribution from the long-horizon difficulty axes.

\paragraph{Historical evidence informing this design.}
On six controlled hidden-root cases, dependency slicing recovered the root in
all six top-three sets while the generic and Who\&When-style controls recovered
only one; on four retrospective natural development trajectories, dependency
candidate construction recalled the verified root in all four and blind
dependency ranking reached Hit@1 on three, whereas a blind full-trace ranker
reached Hit@1 on none. These small development cells are not pooled with the
$N=200$ estimate.

\subsection{RQ2: Can FIX recover better from the same fault information?}
\label{sec:fix-results}

\begin{figure}[t]
    \centering
    \includegraphics[width=\linewidth]{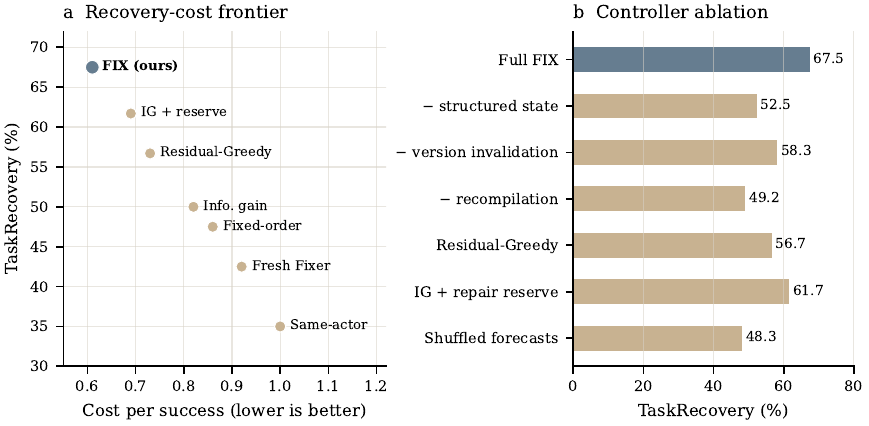}
    \caption{\textbf{FIX recovery, cost, and mechanism.}
    (a) Recovery--cost frontier for the methods in Table~\ref{tab:fix-main-expanded},
    plotting \taskrecovery{} against cost per success; lower cost is better.
    (b) Mechanism ablation under matched information and legal actions.
    The strongest planning control is information gain with a repair reserve,
    not a budget-oblivious strawman.}
    \label{fig:fix-recovery}
\end{figure}

Table~\ref{tab:fix-main-expanded} removes TRACE as a confound by supplying every
method with the same independently verified fault packet, failed state, legal
tools, and intervention budget. The strongest non-DP comparator is
information gain with an explicit repair reserve, because controlled
development showed that this policy can match dynamic planning when the main
difficulty is simply preserving enough budget to finish the repair.
Figure~\ref{fig:fix-recovery} plots recovery against cost, so a controller that
succeeds only by spending more is not confused with one that reaches a better
recovery--cost frontier.

\paragraph{Why these controls are necessary.}
In a 24-case executable controlled verification study, naive adaptive
information gain resolved $8/24$ cases at budget four, while information gain
with repair reservation and dynamic TRACE-DP each resolved $16/24$.
That result says the controlled gain was explained by reserving completion
budget and does \emph{not} establish a unique Bellman-planning advantage.
The natural matched-fault experiment above is therefore the test that can
separate residual-state planning from this strong alternative.

\subsection{RQ3: Do TRACE and FIX compose in a real closed loop?}
\label{sec:closed-loop-results}

Component improvements are insufficient if the TRACE packet does not survive
into executable recovery. We therefore run all four combinations
$T_0F_0,T_0F_1,T_1F_0,T_1F_1$ from the same failed entry state with the same
actor, legal tools, budget, stopping rule, and official verifier.
\begin{figure}[t]
    \centering
    \includegraphics[width=0.9\linewidth]{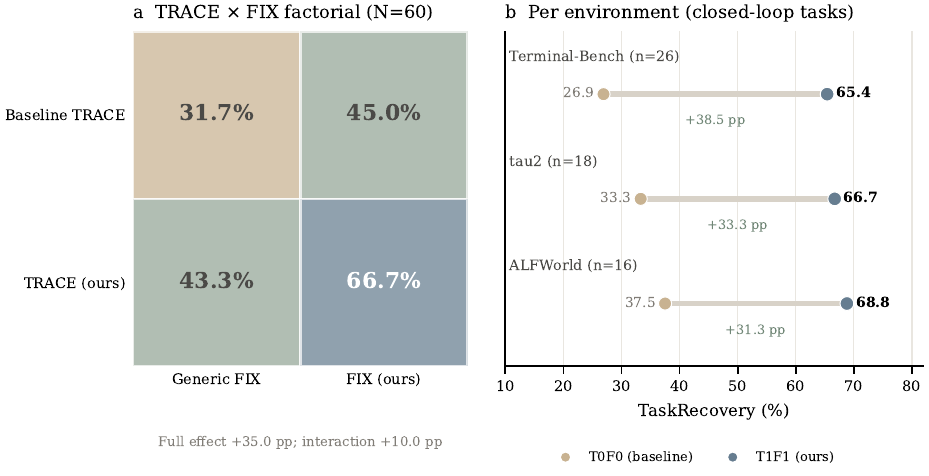}
    \caption{\textbf{From diagnosis to task recovery.}
    (a) The $2\times2$ TRACE$\times$FIX crossing reports benchmark-owned
    \taskrecovery{}. Replacing only TRACE or only FIX isolates each component;
    the diagonal gives the end-to-end system.
    (b) Per-environment \taskrecovery{} for $T_0F_0$ and $T_1F_1$ on the 60
    closed-loop tasks, which span three environments (26, 18 and 16 tasks).}
    \label{fig:closed-loop}
\end{figure}
The matrix is $31.7\%$ for $T_0F_0$, $45.0\%$ for $T_0F_1$,
$43.3\%$ for $T_1F_0$, and $66.7\%$ for $T_1F_1$, corresponding to a
$+35.0$ point full effect and a $+10.0$ point interaction.
The environment split is important because a single aggregate can otherwise
hide a recovery mechanism that works only for one kind of state.

\subsection{RQ4: Which parts of the recovery controller matter?}
\label{sec:ablation-results}

The primary ablations preserve the actor, public information, legal action
catalogue, base forecast source, and intervention budget. Each row removes one
decision-relevant mechanism rather than replacing the method with an unrelated
weak agent.
Three additional diagnostics are required before a
strong planning claim is made. First, we measure check-verdict calibration and
actor-continuation calibration of the public-state forecast model on held-out
executed transitions. Second, we compare the selected action against an
oracle-outcome upper bound and report action-ranking regret rather than treating
forecast probabilities as calibrated by construction.

Third, we report the
fraction of decisions that enter an actionable check/repair path, the share of
executed checks returning \emph{unknown}, and the stage at which each failed
recovery terminates. These diagnostics are motivated by retained runs in which
one controller spent budget on three unknown checks while another issued
51 consecutive actor-continuation decisions without entering recovery.

\begin{table}[t]
\caption{\textbf{Controller ablations under matched verified-fault input.}
The information-gain+reserve row is also a strong
alternative policy, not merely an ablation.}
\label{tab:controller-ablation}
\centering
\scriptsize
\setlength{\tabcolsep}{3.0pt}%
\renewcommand{\arraystretch}{1.08}
\begin{tabular}{lrrr}
\toprule
Controller & \taskrecovery{} & Harmful & Cost/succ. \\
\midrule
\textbf{Full FIX}                    & \textbf{67.5} & \textbf{2.5} & \textbf{0.61} \\
No structured residual state         & 52.5 & 5.0 & 0.78 \\
No version invalidation              & 58.3 & 5.8 & 0.74 \\
No receipt-triggered recompilation   & 49.2 & 6.7 & 0.82 \\
Residual-Greedy                      & 56.7 & 3.3 & 0.73 \\
Information gain + repair reserve    & 61.7 & 3.3 & 0.69 \\
Shuffled outcome forecasts           & 48.3 & 5.8 & 0.85 \\
\bottomrule
\end{tabular}
\end{table}

\paragraph{Experience adaptation.}
The optional cross-case outcome adaptation is not part of the primary
experiment chain. Historical guarded development removed negative transfer but
produced no uplift over the non-memory policy. We therefore do not use this component to support the TRACE or FIX claims.

\section{Limitations}
\label{sec:limitations}

TRACE--FIX has several boundaries. First, the residual state is an approximation: missing dependencies, unobserved side effects, or incorrect resource links can exclude relevant events; dependency links narrow hypotheses without establishing causality. Second, FIX relies on bounded lookahead with uncalibrated forecasts, where truncation and estimation errors can affect action selection. Third, recovery requires reproducible failed states, a frozen check-and-repair catalogue, and an independent verifier, so results may not transfer to open-world settings without replay or known repair actions. Finally, recovery value depends on the continuation actor and cost model; we claim neither optimal control nor unique causal identification.

\section{Conclusion}
\label{sec:conclusion}

Diagnosing and recovering from long-horizon agent failures requires decoupling root-cause localization, budgeted state repair, and end-to-end execution. We formalize this separation through \textsc{Trace-Fix}: \textsc{Trace} represents public history as an explicit residual ledger to prioritize unsatisfied task requirements, while \textsc{Fix} conducts bounded dynamic programming over verified public states and runtime execution receipts. Across matched benchmarks and factorial evaluations, structured residual tracking substantially improves root-cause localization, targeted planning enhances recovery yield under equivalent fault packets, and their closed-loop integration outperforms either component in isolation. By converting historical failure explanations into actionable residual obligations, \textsc{Trace-Fix} provides a principled framework for reliable, budgeted recovery in persistent stateful environments.

\bibliographystyle{iclr2027_conference}
\bibliography{references}

\end{document}